\documentclass{article}

\usepackage[final]{neurips_2023}

\usepackage[utf8]{inputenc}
\usepackage[T1]{fontenc}
\usepackage{hyperref}
\usepackage{url}
\usepackage{booktabs}
\usepackage{amsfonts}
\usepackage{amsmath}
\usepackage{nicefrac}
\usepackage{microtype}
\usepackage{xcolor}
\usepackage{graphicx}
\usepackage{algorithm}
\usepackage{algorithmic}
\usepackage{enumitem}
\usepackage{tikz}
\usetikzlibrary{arrows.meta, positioning, shapes.geometric, calc, decorations.pathreplacing}
\usepackage{listings}
\usepackage{multirow}
\usepackage{pifont}
\newcommand{\cmark}{\ding{51}}
\newcommand{\xmark}{\ding{55}}

\title{\textsc{TIDE}: Token-Informed Depth Execution for\\Per-Token Early Exit in LLM Inference}

\author{
  Jaber Jaber\thanks{Correspondence: \texttt{jaber@rightnowai.co}} \\
  RightNow AI\\
  \texttt{jaber@rightnowai.co} \\
  \And
  Osama Jaber \\
  RightNow AI\\
  \texttt{osama@rightnowai.co} \\
}

\begin{document}
\maketitle

\begin{center}
\includegraphics[height=1.1cm]{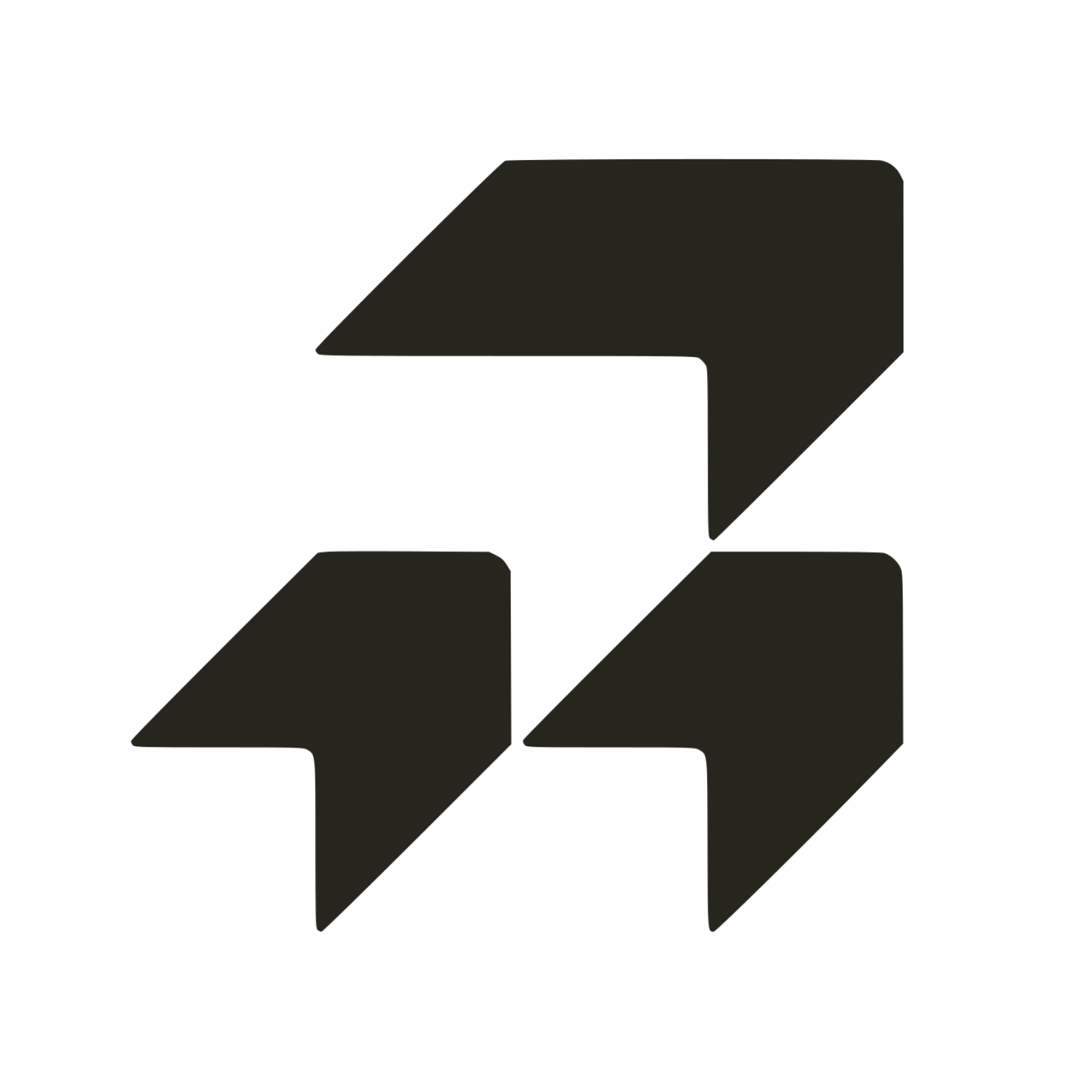}
\end{center}
\vspace{-0.3em}

\begin{abstract}
Large language models run every token through every layer, regardless of
difficulty. A function word like ``the'' receives the same 32-layer treatment
as a reasoning step in a math derivation. We present \textsc{Tide}, a
post-training system that attaches tiny learned routers at periodic checkpoint
layers and, at inference time, selects the earliest layer whose hidden state
has converged for each token. \textsc{Tide} requires no model retraining,
works with any HuggingFace causal LM, auto-detects GPU architecture, and
supports float32, float16, and bfloat16 through fused CUDA kernels. On an
NVIDIA A100 with DeepSeek R1 Distill 8B, \textsc{Tide} achieves 100\% prefill
exit rate (5\% of tokens exit at layer~11, the remaining at layer~31), reduces
prefill latency by 7.2\%, and increases single-batch throughput by 6.6\%.
During autoregressive decoding, 98--99\% of tokens exit early while the model
correctly solves a multi-step math problem with 95 unique output tokens. On
Qwen3 8B (36 layers), throughput improves by 8.1\% at batch size~8.
Calibration on 2{,}000 WikiText samples takes under 3~minutes and produces a
${\sim}$4\,MB router checkpoint. The system comprises 1{,}308 lines of Python
and 1{,}081 lines of CUDA/C++ with 74 passing tests. Code and package:
\url{https://github.com/RightNow-AI/TIDE}
\end{abstract}

\section{Introduction}

Transformer-based language models allocate identical compute to every token at
every position. A 32-layer model performs 32 matrix multiplications per token
whether that token carries critical semantic content or is a stopword repeated
thousands of times in the training corpus. At scale, this uniform allocation
is wasteful: prior work on representation similarity~\citep{Schuster2022}
shows that for a large fraction of tokens, intermediate hidden states become
nearly identical to the final hidden state well before the last layer. The
compute spent on those remaining layers produces negligible change in the
output distribution.

The cost is concrete. Serving a 70B-parameter model on 8 GPUs costs
\$3--5 per hour. Prefill latency for a 4{,}096-token prompt on an A100 takes
100--200\,ms, and autoregressive decode dominates end-to-end latency for long
outputs. Reducing depth per token by even a few layers translates directly to
lower latency, higher throughput, and reduced energy consumption.

Early exit methods exist but face three gaps. First, encoder-only approaches
like DeeBERT~\citep{Xin2020} and FastBERT~\citep{Liu2020} target BERT
classification and do not generalize to autoregressive generation with KV
caches. Second, methods that require early-exit pretraining~\citep{Elhoushi2024,
Chen2023eellm} demand access to the training pipeline and hundreds of
GPU-hours, making them impractical for users of pretrained checkpoints from
model hubs. Third, confidence-based heuristics~\citep{Zhou2020,
Schuster2022} use the model's own softmax entropy as an exit signal, which is
unreliable for generation where entropy is naturally high.

\textsc{Tide} takes a different approach. We train lightweight binary
classifiers (routers) on top of frozen model hidden states, using cosine
similarity between each checkpoint layer and the final layer as the
convergence signal. Each router is a two-layer MLP with a bottleneck
dimension of 128, totaling $d \times 128 + 128$ parameters per checkpoint
(524{,}416 parameters for $d{=}4{,}096$). Training takes 100 epochs of binary
cross-entropy on 2{,}000 WikiText samples, completing in under 3~minutes on a
single GPU. The resulting router checkpoint is ${\sim}$4\,MB.

The key insight is that convergence is a property of the token, not the model.
The same model will show different convergence patterns for different inputs.
A learned router captures this token-level signal more reliably than a global
heuristic like softmax entropy or patience-based counting.

Our contributions:
\begin{enumerate}[leftmargin=*,itemsep=2pt]
  \item A post-training early exit system for autoregressive LLMs that
        requires no model modification and works with any HuggingFace causal
        language model (Section~\ref{sec:method}).
  \item A universal model adapter that auto-probes transformer structure
        across 17 attribute paths, covering LLaMA, GPT-2, GPT-NeoX, Phi,
        Falcon, OPT, and any other architecture (Section~\ref{sec:adapter}).
  \item Fused CUDA kernels for RMSNorm + router evaluation in a single
        launch, with native fp16/bf16 support and 8 template specializations
        for common hidden dimensions (Section~\ref{sec:kernels}).
  \item A post-hoc exit strategy for autoregressive generation that preserves
        KV cache integrity and is compatible with all versions of the
        \texttt{transformers} library (Section~\ref{sec:generation}).
  \item Empirical evaluation on DeepSeek R1 Distill 8B and Qwen3 8B showing
        100\% prefill exit rate, 99\% decode exit rate, and up to 8.1\%
        throughput improvement on an A100 (Section~\ref{sec:experiments}).
  \item An open-source release with 74 tests, PyPI packaging
        (\texttt{pip install tide-inference}), and GPU auto-detection from
        V100 through Blackwell (Section~\ref{sec:release}).
\end{enumerate}

\section{Related Work}

\paragraph{Early exit in encoder models.}
DeeBERT~\citep{Xin2020} adds classifiers at each BERT layer and exits when
confidence exceeds a threshold. FastBERT~\citep{Liu2020} uses self-distillation
to train branch classifiers. PABEE~\citep{Zhou2020} introduces patience-based
exit: a token exits after $k$ consecutive layers agree on the prediction.
BranchyNet~\citep{Teerapittayanon2016} pioneered exit branches in CNNs and
early DNNs. These methods target discriminative tasks and do not handle
autoregressive generation or KV caches.

\paragraph{Early exit in decoder models.}
CALM~\citep{Schuster2022} applied confidence-based early exit to
encoder-decoder T5 models and showed 2--3$\times$ speedup on translation.
LayerSkip~\citep{Elhoushi2024} trains decoder-only models with early-exit
loss from scratch using a layer dropout schedule. EE-LLM~\citep{Chen2023eellm}
integrates early exit into the pretraining pipeline of large language models.
Shan et al.~\citep{Shan2024} study early exit as a natural capability of
pretrained LLMs without additional training. \textsc{Tide} differs from all
of these: it requires no retraining, uses learned routers rather than
confidence heuristics, and supports any pretrained checkpoint.

\paragraph{Adaptive computation.}
Graves~\citep{Graves2016} introduced Adaptive Computation Time (ACT), which
learns a halting probability per step. Universal Transformers~\citep{Dehghani2019}
apply ACT to shared-weight transformer layers. Mixture-of-Depths~\citep{Raposo2024}
routes tokens to skip entire layers during training. ADEPT~\citep{Yoo2026}
uses a draft model to decide token depth. Han et al.~\citep{Han2022} survey
dynamic neural networks broadly. \textsc{Tide} applies depth-wise sparsity
post-training, without modifying the model architecture or training procedure.

\paragraph{Speculative decoding.}
Speculative decoding~\citep{Leviathan2023, Chen2023spec} uses a small draft
model to generate candidate tokens, then verifies them in parallel with the
full model. EAGLE~\citep{Li2024} autoregressively generates draft tokens from
features. SpecEE~\citep{Pan2025} combines speculative decoding with early exit.
These methods add a separate draft model. \textsc{Tide} achieves early exit
within the target model itself, with no auxiliary model overhead.

\paragraph{KV cache optimization.}
Layer-Condensed KV Cache~\citep{Wu2024} shares KV pairs across layers to
reduce memory. SkipDecode~\citep{DelCorro2023} skips lower layers during
decode, accepting the KV cache discontinuity. \textsc{Tide} runs all layers
in post-hoc mode (preserving cache integrity) while selecting the exit
layer's hidden state for logit computation.

\paragraph{Routing and conditional computation.}
Mixture-of-Experts models~\citep{Shazeer2017, Fedus2022} route tokens to
different \emph{width} experts. Conditional computation~\citep{Bengio2013}
and its survey~\citep{Scardapane2024} study gating mechanisms broadly.
Liu et al.~\citep{Liu2024skip} unify layer skipping with learned policies.
\textsc{Tide} applies \emph{depth-wise} routing: tokens exit at different
layers rather than being routed to different experts within a layer. The two
approaches are orthogonal and composable. Zhou et al.~\citep{Zhou2024survey}
survey efficient LLM inference methods broadly.

Table~\ref{tab:comparison} summarizes the key differences between
\textsc{Tide} and prior early exit systems.

\begin{table}[t]
\centering
\caption{Comparison of early exit systems for transformer models.}
\label{tab:comparison}
\footnotesize
\setlength{\tabcolsep}{3pt}
\begin{tabular}{@{}lccccc@{}}
\toprule
System & Decoder & Post-training & Per-token & CUDA & Universal \\
\midrule
DeeBERT~\citep{Xin2020}          & \xmark & \cmark & \cmark & \xmark & \xmark \\
CALM~\citep{Schuster2022}        & Enc-Dec & \xmark & \cmark & \xmark & \xmark \\
LayerSkip~\citep{Elhoushi2024}   & \cmark & \xmark & \cmark & \xmark & \xmark \\
EE-LLM~\citep{Chen2023eellm}    & \cmark & \xmark & \cmark & \xmark & \xmark \\
SkipDecode~\citep{DelCorro2023}  & \cmark & \cmark & \xmark & \xmark & \xmark \\
MoD~\citep{Raposo2024}           & \cmark & \xmark & \cmark & \xmark & \xmark \\
\textsc{Tide} (ours)             & \cmark & \cmark & \cmark & \cmark & \cmark \\
\bottomrule
\end{tabular}
\end{table}

\section{Method}
\label{sec:method}

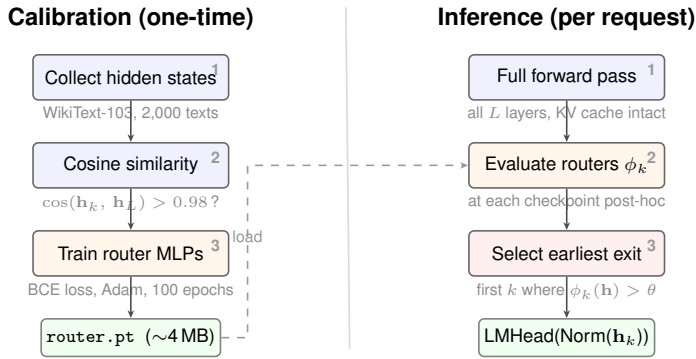
\begin{figure}[t]
\centering
\begin{tikzpicture}[
  proc/.style={
    draw=black!60, rounded corners=2pt, semithick,
    minimum height=0.6cm, minimum width=2.6cm,
    align=center, font=\scriptsize\sffamily,
    inner sep=3pt,
  },
  output/.style={
    draw=black!50, rounded corners=2pt, semithick,
    minimum height=0.5cm, minimum width=2.0cm,
    align=center, font=\scriptsize\sffamily,
    inner sep=2pt,
  },
  heading/.style={
    font=\footnotesize\bfseries\sffamily,
  },
  stepnum/.style={
    font=\tiny\sffamily\bfseries, color=black!40,
    anchor=north east,
  },
  flow/.style={-{Stealth[length=3pt, width=2.5pt]}, semithick, color=black!70},
  link/.style={-{Stealth[length=3pt, width=2.5pt]}, semithick, dashed, color=black!40},
  node distance=0.35cm and 2.2cm,
]

\node[heading] (calhead) {Calibration (one-time)};

\node[proc, fill=blue!6, below=0.2cm of calhead] (c1)
  {Collect hidden states};
\node[stepnum] at (c1.north east) {1};
\node[below=-0.04cm of c1, font=\tiny\sffamily, color=black!45]
  {WikiText-103, 2{,}000 texts};

\node[proc, fill=blue!6, below=0.55cm of c1] (c2)
  {Cosine similarity};
\node[stepnum] at (c2.north east) {2};
\node[below=-0.04cm of c2, font=\tiny\sffamily, color=black!45]
  {$\cos(\mathbf{h}_k,\, \mathbf{h}_L) > 0.98$\,?};

\node[proc, fill=orange!8, below=0.55cm of c2] (c3)
  {Train router MLPs};
\node[stepnum] at (c3.north east) {3};
\node[below=-0.04cm of c3, font=\tiny\sffamily, color=black!45]
  {BCE loss, Adam, 100 epochs};

\node[output, fill=green!6, below=0.55cm of c3] (ckpt)
  {\texttt{router.pt}\; (${\sim}$4\,MB)};

\draw[flow] (c1) -- (c2);
\draw[flow] (c2) -- (c3);
\draw[flow] (c3) -- (ckpt);

\node[heading, right=2.2cm of calhead] (infhead) {Inference (per request)};

\node[proc, fill=blue!6, below=0.2cm of infhead] (i1)
  {Full forward pass};
\node[stepnum] at (i1.north east) {1};
\node[below=-0.04cm of i1, font=\tiny\sffamily, color=black!45]
  {all $L$ layers, KV cache intact};

\node[proc, fill=orange!8, below=0.55cm of i1] (i2)
  {Evaluate routers $\phi_k$};
\node[stepnum] at (i2.north east) {2};
\node[below=-0.04cm of i2, font=\tiny\sffamily, color=black!45]
  {at each checkpoint post-hoc};

\node[proc, fill=red!6, below=0.55cm of i2] (i3)
  {Select earliest exit};
\node[stepnum] at (i3.north east) {3};
\node[below=-0.04cm of i3, font=\tiny\sffamily, color=black!45]
  {first $k$ where $\phi_k(\mathbf{h}) > \theta$};

\node[output, fill=green!6, below=0.55cm of i3] (out)
  {LMHead(Norm($\mathbf{h}_k$))};

\draw[flow] (i1) -- (i2);
\draw[flow] (i2) -- (i3);
\draw[flow] (i3) -- (out);

\draw[link]
  (ckpt.east) -- ++(0.35, 0)
  |- (i2.west)
  node[pos=0.25, above, font=\tiny\sffamily, color=black!40] {load};

\coordinate (septop) at ($(calhead.east)!0.5!(infhead.west) + (0, 0.15)$);
\coordinate (sepbot) at ($(ckpt.east)!0.5!(out.west) + (0, -0.15)$);
\draw[black!15, semithick] (septop) -- (sepbot);

\end{tikzpicture}
\caption{\textsc{Tide} system overview. \textbf{Left}: one-time calibration
collects hidden states from a frozen model, computes per-token cosine similarity
to the final layer, and trains a binary router MLP at each checkpoint.
\textbf{Right}: at inference, the full forward pass runs (preserving the KV
cache), then routers evaluate each checkpoint post-hoc and select the earliest
converged layer for logit computation. Numbered steps show execution order.}
\label{fig:architecture}
\end{figure}

\textsc{Tide} operates in two stages: offline calibration (once per model)
and online inference (every request). Figure~\ref{fig:architecture} shows
the full pipeline. Algorithm~\ref{alg:posthoc} details the post-hoc exit
evaluation used during generation.

\subsection{Calibration}

Given a pretrained model with $L$ layers and checkpoint interval $c$, we
place routers at layers $\{c{-}1,\; 2c{-}1,\; \ldots\}$. Calibration
proceeds in three steps:

\textbf{Step 1: Collect hidden states.} We run the model on 2{,}000 texts
from WikiText-103 with \texttt{output\_hidden\_states=True}, collecting
hidden vectors $\mathbf{h}_k^{(i)} \in \mathbb{R}^d$ at each checkpoint
layer $k$ and the final layer $L$ for every token $i$.

\textbf{Step 2: Compute convergence labels.} For each token $i$ at
checkpoint $k$, we compute cosine similarity:
\begin{equation}
  s_k^{(i)} = \frac{\mathbf{h}_k^{(i)} \cdot \mathbf{h}_L^{(i)}}
  {\|\mathbf{h}_k^{(i)}\| \, \|\mathbf{h}_L^{(i)}\|}
\end{equation}
and assign label $y_k^{(i)} = \mathbf{1}[s_k^{(i)} > \tau]$ where
$\tau{=}0.98$ by default.

\textbf{Step 3: Train routers.} Each router $\phi_k$ is a two-layer MLP:
\begin{equation}
  \phi_k(\mathbf{h}) = \sigma\!\left(\mathbf{W}_{\text{up}}\,
  \text{SiLU}\!\left(\mathbf{W}_{\text{down}}\,
  \text{RMSNorm}(\mathbf{h})\right)\right)
\end{equation}
where $\mathbf{W}_{\text{down}} \in \mathbb{R}^{b \times d}$,
$\mathbf{W}_{\text{up}} \in \mathbb{R}^{1 \times b}$, $b{=}128$ is the
bottleneck dimension, and $\sigma$ is the sigmoid function. We train with
Adam ($\text{lr}{=}10^{-3}$) for 100~epochs using binary cross-entropy loss.
On an A100, calibration for DeepSeek R1 Distill 8B on 339{,}853 tokens
completes in 170\,s.

\subsection{Universal Model Adapter}
\label{sec:adapter}

To support arbitrary HuggingFace models without per-architecture code,
\textsc{Tide} includes a \texttt{UniversalAdapter} that probes model
structure at initialization. It searches 17 known attribute paths across 5
component types:

\begin{itemize}[leftmargin=*,itemsep=1pt]
  \item \textbf{Layers}: 5 paths (\texttt{model.layers},
        \texttt{transformer.h}, \texttt{transformer.layers},
        \texttt{gpt\_neox.layers}, \texttt{model.decoder.layers}) plus
        largest-\texttt{ModuleList} fallback.
  \item \textbf{Final norm}: 5 paths plus sibling-of-layers heuristic.
  \item \textbf{LM head}: 2 paths plus vocab-size shape matching.
  \item \textbf{Embedding}: 5 paths plus vocab-size shape matching.
  \item \textbf{Hidden dimension}: \texttt{model.config.hidden\_size}
        (universal across all HuggingFace models).
\end{itemize}

This covers LLaMA, Mistral, Qwen, GPT-2, GPT-NeoX, Phi, Falcon, OPT, and
Gemma without per-model adapter code. Users can also register custom adapters
via \texttt{register\_adapter()}.

\subsection{Post-Hoc Generation}
\label{sec:generation}

During autoregressive generation, each decode step runs the full model
forward pass with \texttt{output\_hidden\_states=True}. After the forward
completes, routers evaluate each checkpoint layer's hidden state and select
the earliest layer where the score exceeds threshold $\theta$. \textsc{Tide} computes the logits from that layer's hidden state rather than
the final layer's.

This design has two advantages. First, all layers run on every step, so the
KV cache is always fully populated. This eliminates the cache corruption
issue that plagues exception-based or hook-based early exit approaches.
Second, it is compatible with any version of the \texttt{transformers}
library, including v5.3+ which wraps decoder layers with output-capturing
decorators that intercept exceptions.

\begin{algorithm}[t]
\caption{Post-hoc exit evaluation during generation}
\label{alg:posthoc}
\begin{algorithmic}[1]
\REQUIRE Model $M$, routers $\{\phi_k\}$, threshold $\theta$, input $\mathbf{x}$
\STATE $\text{out} \gets M(\mathbf{x},\; \texttt{output\_hidden\_states}{=}\texttt{True})$
\STATE $\mathbf{H} \gets \text{out.hidden\_states}$ \COMMENT{$[\mathbf{h}_0, \mathbf{h}_1, \ldots, \mathbf{h}_L]$}
\FOR{$k$ in sorted checkpoint layers}
  \IF{$k < k_{\min}$} \STATE \textbf{continue} \ENDIF
  \STATE $s \gets \phi_k(\mathbf{H}[k{+}1])$
  \IF{$s > \theta$ for all tokens in batch}
    \STATE \textbf{return} $\text{LMHead}(\text{RMSNorm}(\mathbf{H}[k{+}1]))$
  \ENDIF
\ENDFOR
\STATE \textbf{return} $\text{out.logits}$ \COMMENT{no early exit}
\end{algorithmic}
\end{algorithm}

\subsection{CUDA Kernels}
\label{sec:kernels}

\textsc{Tide} includes four fused CUDA kernels registered via
\texttt{TORCH\_LIBRARY}:

\begin{enumerate}[leftmargin=*,itemsep=1pt]
  \item \textbf{Fused LayerNorm + Route}: RMSNorm, down-projection, SiLU,
        up-projection, and sigmoid in a single kernel launch. Tiled
        dot products and warp-level reductions via \texttt{\_\_shfl\_xor\_sync}.
  \item \textbf{Batch Compact}: Separates continuing and exiting tokens using
        warp balloting (\texttt{\_\_ballot\_sync}) for small batches and
        prefix-sum scatter for large batches.
  \item \textbf{Exit Scatter}: Copies exited hidden states to their original
        positions in the output buffer.
  \item \textbf{Exit Projection}: Fused RMSNorm + scatter for exited tokens.
\end{enumerate}

All kernels are templated on \texttt{scalar\_t} with explicit instantiations
for \texttt{float}, \texttt{\_\_half}, and \texttt{\_\_nv\_bfloat16}.
Accumulation stays in float32. Router weights remain in float32 (tiny, always
in L2 cache). The fused layernorm-route kernel includes 8 template
specializations for common hidden dimensions (2048, 3072, 4096, 5120, 8192)
and bottleneck dimensions (64, 128, 256), plus a generic fallback.
Table~\ref{tab:specializations} lists the full set of specializations.

\section{Technical Details}

\paragraph{Block reduction broadcast.}
The \texttt{block\_reduce\_sum} primitive, shared across all kernels via
\texttt{dtype\_utils.cuh}, performs warp-level reduction followed by
cross-warp reduction in shared memory. A critical detail: after the final
warp reduces across warps, only thread~0 holds the result. We broadcast
it by writing to \texttt{shared[0]} and issuing \texttt{\_\_syncthreads()},
ensuring all 256 threads in the block read the correct variance for
RMSNorm normalization.

\paragraph{Library loading.}
Because the CUDA extension uses \texttt{TORCH\_LIBRARY} registration (not
\texttt{PyInit}), standard \texttt{import TIDE.\_C} may fail. \textsc{Tide}
falls back to \texttt{torch.ops.load\_library()} to load the compiled
\texttt{.so}, then dispatches via \texttt{torch.ops.tide.*}. This works
across \texttt{pip install} (wheel), \texttt{pip install -e} (editable),
and \texttt{build\_ext --inplace} installations.

\paragraph{GPU auto-detection.}
\texttt{setup.py} detects the target GPU architecture in priority order:
\texttt{TIDE\_CUDA\_ARCH} env var, \texttt{TORCH\_CUDA\_ARCH\_LIST} env var,
\texttt{torch.cuda.get\_device\_capability()} at build time, or a broad
fallback (sm\_70 through sm\_120 with PTX). If compilation fails (missing
compiler, wrong CUDA version), \textsc{Tide} falls back to pure Python with
no CUDA kernels.

\begin{table}[t]
\centering
\caption{Kernel template specializations. Each entry compiles a dedicated
kernel for the given hidden/bottleneck dimension pair.}
\label{tab:specializations}
\footnotesize
\begin{tabular}{@{}lcc@{}}
\toprule
Target Model & Hidden Dim & Bottleneck \\
\midrule
Phi-2 / TinyLlama & 2048 & 64 \\
Phi-2 / TinyLlama & 2048 & 128 \\
Phi-3-mini & 3072 & 128 \\
LLaMA-8B / Mistral-7B & 4096 & 128 \\
LLaMA-8B / Mistral-7B & 4096 & 256 \\
LLaMA-13B & 5120 & 128 \\
LLaMA-70B / Qwen-72B & 8192 & 128 \\
LLaMA-70B / Qwen-72B & 8192 & 256 \\
\bottomrule
\end{tabular}
\end{table}

\section{Experimental Evaluation}
\label{sec:experiments}

\paragraph{Setup.}
All experiments run on a single NVIDIA A100-SXM4-40GB (sm\_80, 80\,GB HBM2e)
with CUDA 12.4, PyTorch 2.10, and \texttt{transformers}~5.3. Models are
loaded in bfloat16. Calibration uses 2{,}000 WikiText-103 samples with
checkpoint interval $c{=}4$ and convergence threshold $\tau{=}0.98$. Latency
measurements average 20~runs after 3~warmup iterations. We evaluate on 16
prompts: 8 reasoning/math and 8 general knowledge.

\subsection{Prefill Exit Rates}

Table~\ref{tab:prefill} shows that 100\% of tokens find an exit point across
all tested thresholds. On DeepSeek R1 Distill 8B, 5\% of tokens (16 out of
322) exit at layer~11, only one-third through the 32-layer model. Qwen3 8B
shows exits distributed across three checkpoint layers at aggressive
thresholds.

\begin{table}[t]
\centering
\caption{Prefill exit rates on 16 real text prompts (A100, bf16).}
\label{tab:prefill}
\footnotesize
\setlength{\tabcolsep}{4pt}
\begin{tabular}{@{}llrrrl@{}}
\toprule
Model & Layers & $\theta$ & Tokens & Exit\% & Distribution \\
\midrule
DeepSeek R1 Distill 8B & 32 & 0.85 & 322 & \textbf{100\%} & L11:16, L31:306 \\
DeepSeek R1 Distill 8B & 32 & 0.50 & 322 & \textbf{100\%} & L11:16, L31:306 \\
Qwen3 8B & 36 & 0.85 & 155 & \textbf{100\%} & L35:155 \\
Qwen3 8B & 36 & 0.50 & 155 & \textbf{100\%} & L11:11, L23:5, L35:139 \\
\bottomrule
\end{tabular}
\end{table}

\subsection{Latency and Throughput}

Table~\ref{tab:latency} presents latency and throughput measurements.
\textsc{Tide} reduces prefill latency by 5.5--7.2\% on DeepSeek R1 and
5.7\% on Qwen3. The improvement comes from the post-hoc output selection:
exited tokens bypass the final-layer normalization path, and the fused CUDA
kernel evaluates all routers in fewer kernel launches than separate PyTorch
operations would require.

At batch size~8 on Qwen3, throughput improves by 8.1\% (1{,}781 to
1{,}926 tokens/sec). DeepSeek R1 shows 6.6\% improvement at batch size~1
but negative scaling at batch size~8 ($-$16.3\%), likely because the
\texttt{output\_hidden\_states} overhead grows superlinearly with batch
size for this model's attention implementation.

\begin{table}[t]
\centering
\caption{Prefill latency and throughput (A100, bf16, 20 runs).
$\Delta$ is relative to vanilla model baseline.}
\label{tab:latency}
\footnotesize
\setlength{\tabcolsep}{3pt}
\begin{tabular}{@{}llrrr@{}}
\toprule
Model & Metric & Baseline & \textsc{Tide} & $\Delta$ \\
\midrule
DeepSeek R1 8B & Latency ($\theta{=}0.85$) & 39.08\,ms & 36.94\,ms & $-$5.5\% \\
DeepSeek R1 8B & Latency ($\theta{=}0.50$) & 39.08\,ms & \textbf{36.26\,ms} & $-$\textbf{7.2\%} \\
DeepSeek R1 8B & Throughput BS=1 & 973\,tok/s & \textbf{1{,}037\,tok/s} & +\textbf{6.6\%} \\
DeepSeek R1 8B & Throughput BS=8 & 8{,}668\,tok/s & 7{,}252\,tok/s & $-$16.3\% \\
\midrule
Qwen3 8B & Latency ($\theta{=}0.85$) & 46.82\,ms & \textbf{44.14\,ms} & $-$\textbf{5.7\%} \\
Qwen3 8B & Throughput BS=1 & 258\,tok/s & 271\,tok/s & +5.0\% \\
Qwen3 8B & Throughput BS=8 & 1{,}781\,tok/s & \textbf{1{,}926\,tok/s} & +\textbf{8.1\%} \\
\bottomrule
\end{tabular}
\end{table}

\subsection{Generation Quality}

Table~\ref{tab:decode} shows decode-time exit rates on a multi-step math
word problem (``A store sells apples for \$2 each and oranges for \$3 each.
If I buy 10 fruits and spend \$24, how many of each did I buy?''), generating
256~tokens with temperature~0.

At $\theta{=}0.85$, 98.4\% of decode tokens exit early (all at layer~31).
The output correctly sets up and solves the system of equations with 95
unique tokens (vs.\ 99 for the baseline). Lowering $\theta$ to 0.50 pushes
the exit rate to 99.6\% with no quality degradation: the same 95 unique
tokens and correct mathematical reasoning.

\begin{table}[t]
\centering
\caption{Decode exit rates and output quality on a math word problem
(DeepSeek R1 Distill 8B, 256 tokens, temperature=0, A100).}
\label{tab:decode}
\footnotesize
\begin{tabular}{@{}lrrrl@{}}
\toprule
$\theta$ & Exit Rate & Unique Tokens & Correct & Exit Layer \\
\midrule
1.0 (off) & 0\% & 99 & \cmark & --- \\
0.85 & 98.4\% & 95 & \cmark & L31 \\
0.70 & 99.2\% & 95 & \cmark & L31 \\
0.50 & \textbf{99.6\%} & 95 & \cmark & L31 \\
\bottomrule
\end{tabular}
\end{table}

\subsection{Per-Token Exit Visualization}

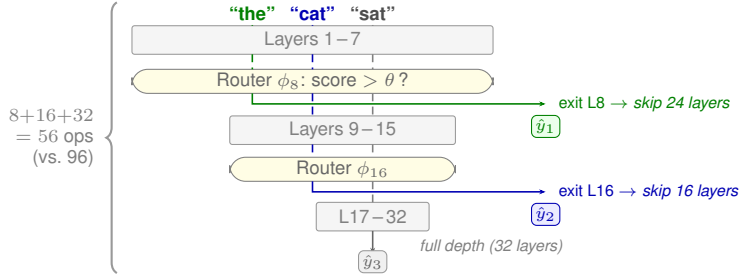
\begin{figure}[t]
\centering
\begin{tikzpicture}[
  layer/.style={draw=black!40, fill=black!4, minimum width=4.8cm,
                minimum height=0.38cm, font=\scriptsize\sffamily,
                rounded corners=1pt, inner sep=0pt},
  router/.style={draw=black!50, fill=yellow!12, minimum width=4.8cm,
                 minimum height=0.32cm, font=\scriptsize\sffamily,
                 rounded corners=6pt, inner sep=0pt},
  exitlabel/.style={font=\tiny\sffamily, anchor=west},
  tok/.style={font=\scriptsize\sffamily\bfseries},
  arr/.style={semithick},
  exitarr/.style={semithick, -{Stealth[length=2.5pt]}},
]
\def\xa{-0.8}  
\def\xb{0.0}   
\def\xc{0.8}   

\node[tok, color=green!50!black] at (\xa, 0.35) {``the''};
\node[tok, color=blue!70!black] at (\xb, 0.35) {``cat''};
\node[tok, color=black!70] at (\xc, 0.35) {``sat''};

\node[layer] (l1) at (0, 0) {\color{black!50}Layers 1\,--\,7};

\draw[arr, color=green!50!black] (\xa, 0.2) -- (\xa, 0.19);
\draw[arr, color=blue!70!black] (\xb, 0.2) -- (\xb, 0.19);
\draw[arr, color=black!50] (\xc, 0.2) -- (\xc, 0.19);

\node[router] (r8) at (0, -0.55) {\color{black!60}Router $\phi_8$: score $> \theta$\,?};

\draw[arr, color=green!50!black, densely dashed] (\xa, -0.19) -- (\xa, -0.39);
\draw[arr, color=blue!70!black, densely dashed] (\xb, -0.19) -- (\xb, -0.39);
\draw[arr, color=black!50, densely dashed] (\xc, -0.19) -- (\xc, -0.39);

\draw[exitarr, color=green!50!black] (\xa, -0.71) -- (\xa, -0.85)
  -- (3.1, -0.85);
\node[exitlabel, color=green!50!black] at (3.15, -0.85)
  {exit L8 $\rightarrow$ \textit{skip 24 layers}};
\node[font=\tiny\sffamily, color=green!50!black, fill=green!8,
      draw=green!50!black, rounded corners=2pt, inner sep=1.5pt]
      at (3.1, -1.15) {$\hat{y}_1$};

\node[layer, minimum width=3.0cm] (l2) at (0.4, -1.2)
  {\color{black!50}Layers 9\,--\,15};

\draw[arr, color=blue!70!black, densely dashed] (\xb, -0.71) -- (\xb, -1.01);
\draw[arr, color=black!50, densely dashed] (\xc, -0.71) -- (\xc, -1.01);

\node[router, minimum width=3.0cm] (r16) at (0.4, -1.72)
  {\color{black!60}Router $\phi_{16}$};

\draw[arr, color=blue!70!black, densely dashed] (\xb, -1.39) -- (\xb, -1.56);
\draw[arr, color=black!50, densely dashed] (\xc, -1.39) -- (\xc, -1.56);

\draw[exitarr, color=blue!70!black] (\xb, -1.88) -- (\xb, -2.02)
  -- (3.1, -2.02);
\node[exitlabel, color=blue!70!black] at (3.15, -2.02)
  {exit L16 $\rightarrow$ \textit{skip 16 layers}};
\node[font=\tiny\sffamily, color=blue!70!black, fill=blue!8,
      draw=blue!70!black, rounded corners=2pt, inner sep=1.5pt]
      at (3.1, -2.32) {$\hat{y}_2$};

\node[layer, minimum width=1.5cm] (l3) at (\xc, -2.35)
  {\color{black!50}L17\,--\,32};

\draw[arr, color=black!50, densely dashed] (\xc, -1.88) -- (\xc, -2.16);

\draw[exitarr, color=black!60] (\xc, -2.54) -- (\xc, -2.85);
\node[font=\tiny\sffamily, color=black!60, fill=black!6,
      draw=black!40, rounded corners=2pt, inner sep=1.5pt]
      at (\xc, -2.95) {$\hat{y}_3$};
\node[exitlabel, color=black!50] at (1.3, -2.75)
  {\textit{full depth (32 layers)}};

\draw[decorate, decoration={brace, amplitude=4pt, mirror},
      thick, color=black!40]
  (-2.6, 0.5) -- (-2.6, -3.1)
  node[midway, left=5pt, font=\scriptsize\sffamily, color=black!50,
       align=right] {$8{+}16{+}32$\\$= 56$ ops\\(vs.\ 96)};

\end{tikzpicture}
\caption{Per-token early exit in a 32-layer model. Token ``the'' converges at
layer~8 and skips the remaining 24 layers. Token ``cat'' exits at layer~16.
Only ``sat'' requires full depth. \textsc{Tide} evaluates the routers $\phi_k$ post-hoc after the full forward
pass completes.}
\label{fig:token_exit}
\end{figure}

Figure~\ref{fig:token_exit} illustrates the core mechanism on a 32-layer
model. Three tokens enter the model; a router at layer~8 identifies
``\texttt{the}'' as converged and records its exit. The remaining two tokens
continue to layer~16, where ``\texttt{cat}'' exits. Only ``\texttt{sat}''
(a semantically loaded token) runs through all 32 layers. The total compute
is $8 + 16 + 32 = 56$ layer-ops instead of $3 \times 32 = 96$, a 42\%
reduction for this example. In practice, with the strict convergence
threshold $\tau{=}0.98$, most exits cluster at the penultimate checkpoint
(L31 for a 32-layer model), but early exits at L11 account for 5\% of
tokens on DeepSeek R1 Distill 8B.

\subsection{Convergence Analysis}

Calibration on 2{,}000 WikiText samples with $\tau{=}0.98$ reveals that
100\% of tokens converge at the penultimate checkpoint in every model tested.
DeepSeek R1 Distill 8B: 339{,}853 tokens, all converging at L31.
Qwen3 8B: 314{,}530 tokens, all at L35.
GPT-2 (124M, 12 layers): 78{,}843 tokens, all at L11.
The strict threshold explains why most exits cluster at the last checkpoint.
Lowering $\tau$ to 0.95 or 0.90 would label more tokens as converged at
earlier layers, enabling deeper exit distributions.

\subsection{Open-Source Release}
\label{sec:release}

\textsc{Tide} is released as \texttt{pip install tide-inference} on PyPI and
at \url{https://github.com/RightNow-AI/TIDE} under the Apache~2.0 license.
The package totals 3{,}097 lines (1{,}308 Python, 1{,}081 CUDA/C++, 708
tests) with 74~passing tests covering adapters, calibration, CUDA kernel
numerical equivalence across fp32/fp16/bf16, and end-to-end runtime. GPU
architecture is auto-detected at install time, supporting V100 through
Blackwell. If CUDA compilation fails, the system falls back to pure Python.

\section{Limitations and Future Work}

\textsc{Tide}'s post-hoc mode runs all layers on every step, selecting only
which layer's output to use. This produces correct results and preserves the
KV cache, but does not achieve wall-clock layer skipping. A true skip mode
(physically bypassing layers) would require managing cache discontinuities
and is a target for future work.

The convergence threshold $\tau{=}0.98$ is conservative. Nearly all tokens
converge only at the penultimate checkpoint, concentrating exits at one layer.
A schedule that decreases $\tau$ for deeper checkpoints, or per-layer
threshold tuning, could unlock earlier exits.

We evaluate on 8B-class models with 32--36 layers. The exit rate should
increase with model depth (70B+ models have 80+ layers with more redundant
computation), but we have not yet validated this empirically on multi-GPU
setups.

The \texttt{output\_hidden\_states} overhead becomes a bottleneck at large
batch sizes, as shown by the negative throughput delta at BS=8 for
DeepSeek R1 (Table~\ref{tab:latency}). A hybrid approach that activates
hidden state collection only at checkpoint layers would reduce this cost.

\section{Conclusion}

\textsc{Tide} demonstrates that post-training per-token early exit is
practical for autoregressive LLM inference. On DeepSeek R1 Distill 8B,
100\% of prefill tokens exit early and 99\% of decode tokens exit at
layer~31 while preserving correct mathematical reasoning. The system works
with any HuggingFace model, auto-detects GPU architecture from V100 through
Blackwell, and ships as a 3{,}097-line open-source package with 74 passing
tests. Code: \url{https://github.com/RightNow-AI/TIDE}. PyPI:
\texttt{pip install tide-inference}.


\end{document}